 \documentclass[final,5p,times,twocolumn,authoryear]{elsarticle}

\usepackage{amssymb}
\usepackage{lipsum}
\usepackage{amsmath}
\usepackage{booktabs}
\usepackage{tikz}
\usetikzlibrary{positioning, arrows.meta}
\usetikzlibrary{shapes, positioning}

\usepackage{xcolor}

\begin{document}

\begin{frontmatter}

\title{From Generative AI to Innovative AI: An Evolutionary Roadmap}

\author[first]{Seyed Mahmoud Sajjadi Mohammadabadi}

\affiliation[first]{organization={Department of Computer Science and Engineering, University of Nevada, Reno},
            city={Reno},
            state={Nevada},
            country={USA}}

\begin{abstract}

This paper explores the critical transition from Generative Artificial Intelligence (GenAI) to Innovative Artificial Intelligence (InAI). While recent advancements in GenAI have enabled systems to produce high-quality content across various domains, these models often lack the capacity for true innovation. In this context, innovation is defined as the ability to generate novel and useful outputs that go beyond mere replication of learned data. The paper examines this shift and proposes a roadmap for developing AI systems that can not only generate content but also engage in autonomous problem-solving and creative ideation. The work provides both theoretical insights and practical strategies for advancing AI to a stage where it can genuinely innovate, contributing meaningfully to science, technology, and the arts.


\end{abstract}


\end{frontmatter}

\section{Introduction}
\label{sec:introduction}

Artificial intelligence (AI) has made remarkable strides in recent years, with Generative AI (GenAI) leading the charge in transforming content creation across various domains. Models such as GPT-4 \citep{achiam2023gpt} and DALL-E 2 \citep{marcus2022very} have demonstrated impressive capabilities in producing text, images, and other media with unprecedented quality. However, despite these advancements, current generative models primarily operate by recombining existing data rather than generating truly novel ideas \citep{boden2004creative}. The fundamental limitation lies in their reliance on pattern replication rather than autonomous problem-solving and innovation.

This paper explores the transition from GenAI to Innovative AI (InAI), a paradigm that aims to develop AI systems capable of genuine creativity, autonomous ideation, and transformative innovation. By integrating techniques from computational creativity, reinforcement learning, and multimodal reasoning, we propose a roadmap for advancing AI beyond mere content generation to true innovation. This work provides both theoretical insights and practical strategies for designing AI systems that can redefine problems, synthesize cross-domain knowledge, and contribute meaningfully to science, technology, and the arts.


\begin{figure*}[ht]
  \centering
  \begin{tikzpicture}[node distance=2.5cm, auto, every node/.style={font=\small}]
    \node [draw, rectangle, fill=blue!10, rounded corners, text width=4cm, align=center] (genai) {%
      \textbf{Generative AI}\\[4pt]
      - Content Generation\\
      - Data Recombination\\
      - Limited Novelty};
      
    \node [draw, rectangle, fill=green!10, rounded corners, right of=genai, xshift=6cm, text width=4cm, align=center] (inai) {%
      \textbf{Innovative AI}\\[4pt]
      - Autonomous Problem Formulation\\
      - Cross-Domain Synthesis\\
      - Ethical \& Trustworthy\\
      - True Innovation};
    
    \node [draw, ellipse, fill=orange!10, below of=genai, yshift=-1cm, text width=5cm, align=center] (roadmap) {%
      \textbf{Evolutionary Roadmap}\\[3pt]
      \footnotesize (Integrating Reinforcement Learning, Meta-Learning, Multimodal Reasoning, and Human-AI Collaboration)};
    
    \draw[->, thick] (genai) -- node[above, align=center]{Transition \\ Roadmap} (inai);
    \draw[->, thick] (genai) -- (roadmap);
    \draw[->, thick] (roadmap) -- (inai);
  \end{tikzpicture}
  \caption{Conceptual roadmap from GenAI to InAI, illustrating the transition from systems that generate content through data recombination to those capable of autonomous, creative innovation.}
  \label{fig:roadmap}
\end{figure*}
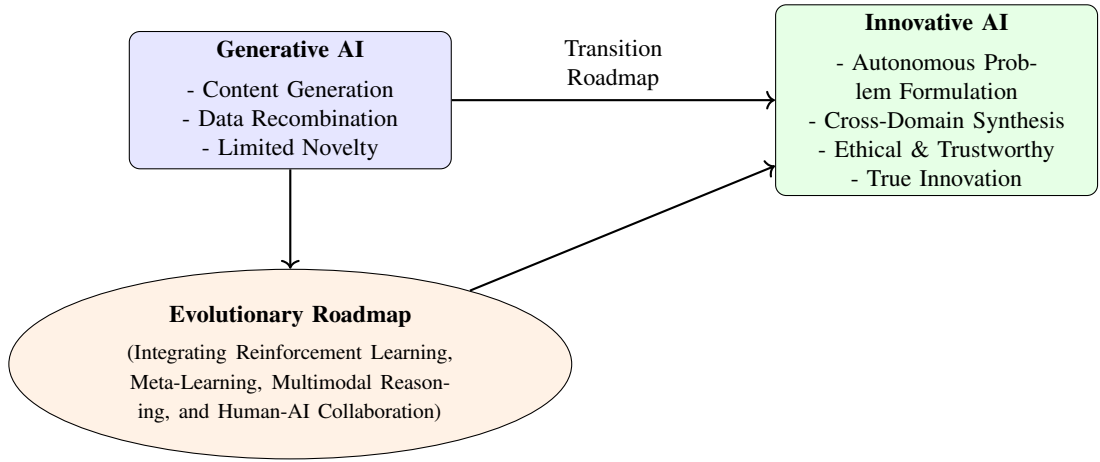


\section{Background}
\label{sec:background}

GenAI techniques have achieved remarkable milestones in creating realistic images, text, and sounds. However, their capacity for genuine innovation is limited. In contrast, computational creativity research \cite{boden2004creative} explores the mechanisms behind human creativity, while reinforcement learning \cite{wiering2012reinforcement} offers methods for continuous improvement based on feedback. These research areas provide essential insights for our proposed model. As shown in Figure \ref{fig:roadmap}, the transition from Generative AI to InAI involves addressing key limitations and integrating advanced learning strategies.


\section{Generative AI} \label{sec:genai}

GenAI has rapidly transformed numerous fields by enabling machines to generate text, images, music, code, and other creative outputs. Leveraging advanced architectures such as transformers \citep{vaswani2017attention}, variational autoencoders (VAEs), and generative adversarial networks (GANs), GenAI models like GPT-4, DALL-E, and Stable Diffusion can produce highly sophisticated content with remarkable efficiency \citep{ramesh2022hierarchical}. Despite these advancements, current generative models remain constrained by several fundamental limitations that hinder their ability to achieve true innovation and autonomous creativity.
Beyond traditional media generation, GenAI has been successfully applied in specialized domains such as smart grid communication. For instance, Mohammadabadi et al. \citep{mohammadabadi2024generative} introduced a distributed learning approach where pre-trained foundation models are fine-tuned locally to generate synthetic data, effectively mitigating communication bottlenecks in smart grids.

\subsection{Limitations of Current Generative AI}

Although GenAI excels at recombining existing data to generate novel outputs, it lacks several essential capabilities required for autonomous innovation, problem-solving, and genuine creativity. Key limitations include:

\subsubsection{Inability to Redefine Problems Autonomously}
One of the fundamental shortcomings of GenAI is its reliance on predefined prompts and user guidance. Unlike human innovators who can redefine problem spaces, challenge assumptions, and generate entirely new research questions, current GenAI models operate within the constraints of their training data and do not possess intrinsic curiosity or autonomous goal-setting capabilities \citep{verganti2020innovation}. True innovation often emerges from questioning established paradigms, an area where GenAI remains deficient.

\subsubsection{Dependence on Training Data and Lack of  Originality}
While GenAI can produce impressive outputs, it fundamentally operates by interpolating and recombining patterns from vast amounts of pre-existing data. This dependence means that it struggles to create genuinely original concepts that do not resemble prior human-generated content. Consequently, GenAI lacks the ability to develop groundbreaking scientific theories, artistic styles, or entirely novel design paradigms that deviate significantly from known structures \citep{boden2004creative}.

\subsubsection{Limited Contextual Understanding and Reasoning}
Although models like GPT-4 exhibit advanced language capabilities, they still struggle with deep reasoning, logical consistency, and long-term coherence in complex domains. GenAI often generates plausible-sounding but factually incorrect or logically inconsistent outputs, highlighting its lack of genuine comprehension \citep{marcus2020next}. This limitation poses significant challenges in domains that require rigorous analytical thinking, such as scientific research, legal reasoning, and strategic decision-making.

\subsubsection{Lack of Multimodal Integration and Cross-Domain Synthesis}
While progress has been made in multimodal AI, current GenAI models primarily function within specific data modalities (e.g., text, images, or audio) rather than seamlessly integrating insights across multiple domains. Human creativity often thrives at the intersection of disciplines, where seemingly unrelated ideas converge to form groundbreaking innovations. Future AI systems must develop more advanced multimodal reasoning capabilities to foster true cross-domain synthesis and innovation \citep{radanliev2024artificial}.

\subsubsection{Challenges in Ethical and Trustworthy AI Generation}
As GenAI becomes more powerful, ethical concerns such as bias, misinformation, and content authenticity become increasingly pressing. AI-generated content can inadvertently reinforce societal biases present in training data, generate deceptive deepfakes, or produce misleading scientific conclusions \citep{bender2021dangers}. Ensuring ethical AI development requires robust mechanisms for bias detection, accountability, and transparent governance frameworks.

\subsection{Toward More Innovative AI Systems}
To move beyond the limitations of current GenAI and towards truly InAI, future research should focus on:
\begin{itemize}
\item \textbf{Developing AI with Autonomous Problem Formulation:} Integrating techniques from reinforcement learning and meta-learning to enable AI to redefine problem spaces rather than merely optimizing within predefined constraints.
\item \textbf{Enhancing Multimodal Reasoning:} Building AI models capable of integrating and reasoning across diverse modalities, unlocking new avenues for interdisciplinary creativity.
\item \textbf{Fostering AI-Driven Cross-Domain Innovation:} Encouraging AI to draw unexpected connections between disparate fields, facilitating breakthroughs in science, engineering, and the arts.
\item \textbf{Improving AI Transparency and Ethical Safeguards:} Implementing proactive mechanisms to mitigate biases, enhance explainability, and ensure AI-generated content aligns with societal values.
\end{itemize}

While current GenAI systems represent a major milestone in artificial intelligence, they remain tools for augmenting human creativity rather than independent innovators. Addressing these limitations will be crucial in transitioning from GenAI to more advanced forms of autonomous and InAI.

\section{Innovative AI}
\label{sec:inai}

InAI represents the next step in the evolution of artificial intelligence. While GenAI systems excel at recombining existing data to produce high-quality outputs, the essence of true innovation and creativity lies in developing and implementing new ideas, methods, or processes that are novel and useful.
\textbf{Innovation} is generally understood as generating and applying ideas that result in significant, positive change. Based on the literature, the key criteria of innovation include:
\begin{itemize}
    \item \textbf{Creativity:} The capacity to produce original ideas that challenge existing paradigms. AI can augment human creativity by expanding the search space for potential solutions, thereby exploring new territories of thought \citep{boden2004creative}.
    \item \textbf{Novelty and Usefulness:} For an idea to be considered innovative, it must be both new and practically applicable. This dual criterion ensures that the idea is not only original but also feasible and appropriate for solving real-world problems \citep{mukherjee2023managing}.
    \item \textbf{Impact:} Innovation should lead to transformative outcomes that yield significant improvements or solve major challenges. Its value is often measured by the degree to which it drives change and creates positive outcomes \citep{hutchinson2020reinventing}.
    \item \textbf{Ethical Considerations:} Responsible innovation involves addressing potential biases, privacy concerns, and other ethical issues to ensure that AI technologies contribute positively to society \citep{nersessian2020automation}.
    \item \textbf{Value Generation:} Beyond novelty, an innovative idea must generate value by enhancing efficiency, unlocking new capabilities, or meeting unmet needs. This value can be both immediate and sustainable over the long term \cite{criado2019creating}.
    \item \textbf{Feasibility and Viability:} A promising innovation must be implementable and viable in practice, meaning that it is technically feasible and capable of being successfully deployed \cite{brenner2022management}.
\end{itemize}

\subsection{Recent Advancements in AI Creativity and Innovation}

Recent advancements in AI creativity and innovation have been significant, particularly with the development of powerful GenAI models:

\subsubsection{Development of Powerful Generative Models}
The launch of models such as ChatGPT (versions 3 and 4), Bard, Stable Diffusion, and DALL-E marks a significant milestone in GenAI. These models, built on architectures like transformers (e.g., GPT), variational autoencoders, and generative adversarial networks (GANs), can perform a wide range of tasks, including text generation, music composition, image creation, video production, and code generation \cite{bengesi2024advancements}. Their capabilities have expanded dramatically, enabling more natural interactions and original content generation \cite{radanliev2024artificial}.

\subsubsection{AI as a Creative Partner and Enabler}
GenAI is increasingly viewed as a collaborative partner in the creative process \cite{sedkaoui2024generative}. Research suggests that working with AI on complex creative tasks enhances an individual’s creative performance and self-efficacy \cite{haase2023artificial}. GenAI acts as a creative catalyst by offering unexpected combinations and challenging existing paradigms, pushing the boundaries of human creativity \cite{sedkaoui2024generative}.

\subsubsection{Idea Generation and Problem Solving}
A key advancement is GenAI’s capability in idea generation and creative problem-solving. These models can autonomously produce diverse content, aligning with creative problem-solving strategies that encourage divergent thinking \cite{sedkaoui2024generative}. Large Language Models (LLMs) are being integrated into the ideation phase, generating original and useful ideas when prompted appropriately, even with minimal examples \cite{bouschery2023augmenting}.

\subsubsection{Overcoming Human Limitations in Innovation}
AI plays a crucial role in overcoming human information-processing constraints. AI systems excel at analyzing vast amounts of data, identifying novel areas for investigation, and suggesting innovative solutions beyond human reach. Advances such as reinforcement learning, particularly unsupervised and meta-reinforcement learning, show promise in enabling algorithms to recognize and achieve goals autonomously, fostering innovation in previously unrelated domains \cite{haefner2021artificial}.

\subsubsection{Impact on Design and Innovation Processes}
The emergence of GenAI necessitates a paradigm shift in conventional innovation approaches, particularly in design thinking. Traditional design thinking models must evolve to leverage AI’s unique capabilities in generating novel and practical ideas. AI also introduces automation into problem-solving, reducing the human-intensive nature of detailed design choices \cite{verganti2020innovation}.

\subsubsection{The Pragmatic Perspective on AI Creativity}
While debates persist on whether AI can be "genuinely" creative, a pragmatic perspective suggests that GenAI can generate original and useful ideas on an everyday level. The focus is shifting towards the perceptual creativity of AI rather than replicating human cognitive processes. The term "Artificial Creativity" has been proposed to describe computer-generated outputs that are both original and effective, even if they lack essential human creative qualities \cite{haase2023artificial}.

\subsubsection{Augmented Creativity: Human-AI Collaboration}
The most optimistic view for the future involves augmented creativity, where human and artificial competencies merge. Humans remain crucial for framing problems, evaluating idea relevance, and implementing solutions, while AI serves as an assistant in provoking, listing, and refining ideas. By leveraging its vast knowledge base, AI expands human thinking and enhances creative processes \cite{haase2023artificial}.

\subsubsection{Applications Across Various Domains}
These advancements have led to practical applications in multiple fields. Language models are being used to generate ideas for patent applications, identify similar patents, monitor communication trends, predict emerging trends, and generate product names. AI is also accelerating innovation in material science, optimizing battery components, discovering new catalysts, and expediting pharmaceutical research and development.

\subsection{Challenges and Gaps in Innovative AI}

Despite significant advancements in GenAI, there are still crucial gaps that must be addressed to realize truly InAI systems. The current landscape of AI creativity and innovation is limited by several key factors:

\subsubsection{Lack of Autonomous Problem Exploration}
One of the fundamental limitations of current AI models is their inability to redefine problems autonomously. While AI excels at optimizing solutions within predefined constraints, true innovation requires the capability to explore and redefine both the problem space and the solution space. Current generative models operate within predefined prompts and user guidance, lacking intrinsic curiosity and exploratory capabilities that could lead to groundbreaking discoveries \cite{verganti2020innovation}.

\subsubsection{Limited Multimodal Integration and Comprehension}
Although AI has made progress in processing multimodal data, existing systems are primarily optimized for specific modalities (e.g., text, image, or video) rather than seamlessly integrating multiple modes of communication. Future AI systems must develop a more profound multimodal understanding, enabling seamless integration across text, images, audio, video, and 3D environments. This would facilitate richer and more InAI-generated content that better mimics human creativity \cite{radanliev2024artificial}.

\subsubsection{Lack of Open-Ended Innovation Mechanisms}
Innovation often arises from unexpected connections across diverse knowledge domains. Current AI models lack mechanisms for open-ended idea generation beyond existing patterns and training data. To achieve true innovation, AI must develop capabilities akin to human serendipity—making unexpected and valuable associations across seemingly unrelated fields \cite{truong2022artificial}.

\subsubsection{Challenges in Steering and Control of Generative AI}
Ensuring that AI-generated content aligns precisely with user intentions remains a significant challenge. Techniques such as dynamic prompting, constrained optimization, and reinforcement learning are promising, yet further advancements are needed to make AI outputs more predictable, controllable, and reliable. AI needs mechanisms for adaptive steering that allow users to refine creative outputs iteratively without starting from scratch.

\subsubsection{Ethical and Trustworthiness Considerations}
As AI becomes increasingly capable of producing human-like creative outputs, the need for effective mechanisms to detect and verify AI-generated content grows. Ethical challenges, including deepfakes, misinformation, and biases, must be addressed through proactive forecasting and ethical frameworks. Future research should explore methods to ensure AI-generated innovations remain aligned with ethical and societal norms.

\section{Possible Directions}
\label{sec:future}

Building on these gaps, future AI research should focus on the following directions to enhance AI-driven innovation:

\subsubsection{Autonomous AI Creativity and Problem-Solving}
Future AI systems should integrate techniques from reinforcement learning, meta-learning, and unsupervised learning to develop the ability to autonomously define, refine, and explore problem spaces. This would enable AI to move beyond merely generating content to becoming an active agent in the creative and innovation process.

\subsubsection{Enhanced Multimodal AI Systems}
GenAI models of the future should seamlessly integrate text, images, audio, video, and 3D environments. Training on diverse multimodal datasets will allow AI to develop a more holistic understanding of creativity and innovation, unlocking new possibilities for content generation across various fields such as design, entertainment, and research.

\subsubsection{Developing AI for Cross-Domain Innovation}
Future research should explore how AI can be trained to identify and generate novel ideas across disparate fields. This could involve designing AI architectures capable of transferring knowledge between unrelated domains, leading to breakthroughs in science, medicine, engineering, and the arts.

\subsubsection{Improving AI Steering and Controllability}
To make AI more manageable, future developments must incorporate more sophisticated methods for user-guided AI output refinement. Techniques such as interactive reinforcement learning, fine-tuned optimization constraints, and dynamic steering should be integrated into generative models to ensure high-quality, relevant, and controllable outputs.

\subsubsection{Ethical AI Development and Trustworthy AI Systems}
Future AI research should prioritize the development of ethical frameworks that proactively address potential ethical challenges associated with AI-generated content. This includes implementing robust detection mechanisms for AI-generated content, preventing deepfake misuse, and ensuring fair and unbiased AI-generated innovations.

\subsubsection{AI-Driven Innovation in Organizations and Research}
Further research should investigate the impact of AI adoption on organizational innovation, considering factors such as firm size, internal vs. external AI integration, and collaboration strategies. Case studies on AI development in various industries, including family-run businesses and tech enterprises, could shed light on best practices for AI-enabled innovation.

\subsubsection{Embedding AI into Strategic Foresight and Innovation Management}
The integration of AI into innovation management and strategic foresight will require advanced data mining models capable of analyzing vast amounts of structured and unstructured data. Future research should explore AI-powered predictive analytics to anticipate emerging trends, technological shifts, and market disruptions.

\subsubsection{Transdisciplinary Approaches to AI-Enabled Innovation}
To fully unlock AI’s innovative potential, interdisciplinary research combining AI with digital humanities, social sciences, and ethics must be encouraged. This would help address societal challenges associated with AI-generated content and foster responsible and inclusive innovation.

\section{Conclusion}
\label{sec:conl}

GenAI has significantly advanced the field of artificial intelligence by enabling the automated creation of high-quality content. However, its reliance on learned patterns and data recombination limits its ability to achieve true innovation. The transition from GenAI to InAI represents the next frontier in AI research, focusing on developing systems that can autonomously redefine problems, synthesize knowledge across domains, and generate novel and useful solutions.
To achieve this, AI must integrate principles from computational creativity, reinforcement learning, and multimodal reasoning. Key areas of future research include enhancing AI’s capacity for autonomous problem formulation, fostering cross-disciplinary synthesis, and ensuring ethical AI development. By addressing these challenges, we can move toward AI systems that not only augment human creativity but also contribute independently to scientific discoveries, technological advancements, and artistic expression.
Ultimately, the shift from generative to InAI will define the next era of artificial intelligence, unlocking new possibilities for autonomous creativity and problem-solving in ways that were previously unimaginable.

\bibliographystyle{elsarticle-harv} 
\bibliography{main}

\end{document}